\documentclass[runningheads]{llncs}

\usepackage{eccv}

\usepackage{eccvabbrv}

\usepackage{graphicx}
\usepackage{booktabs}
\usepackage{multirow}
\usepackage{xcolor}
\usepackage{graphicx}
\usepackage{caption}

\usepackage[accsupp]{axessibility}  

\usepackage[pagebackref,breaklinks,colorlinks,citecolor=eccvblue]{hyperref}

\usepackage{orcidlink}

\graphicspath{{./Figs/}}

\begin{document}

\title{Amped: Adaptive Multi-stage Non-edge Pruning for Edge Detection} 

\titlerunning{Amped}

\author{Yuhan Gao\inst{1,2}\orcidlink{0009-0002-2548-4665} \and
Xinqing Li\inst{1}\orcidlink{0009-0001-0293-4144} \and
Xin He\inst{3}\orcidlink{0009-0004-2139-6590} \and
Bing Li\inst{4}\orcidlink{0000-0002-1875-2919} \and
Xinzhong Zhu\inst{5}\orcidlink{0000-0002-0033-5260} \and
Ming-Ming Cheng\inst{1,2,6}\orcidlink{0000-0001-5550-8758} \and
Yun Liu\inst{1,2,6}\thanks{Corresponding author: Yun Liu (liuyun@nankai.edu.cn)}\orcidlink{0000-0001-6143-0264}}

\authorrunning{Gao et al.}


\institute{VCIP, College of Computer Science, Nankai University \and
Academy for Advanced Interdisciplinary Studies, Nankai University \and
School of Computer Science and Engineering, Tianjin University of Technology \and
School of Information and Communication Engineering, UESTC \and
School of Computer Science and Technology, Zhejiang Normal University \and
Nankai International Advanced Research Institute, Shenzhen Futian}

\maketitle

\begin{abstract}
Edge detection is a fundamental image analysis task that underpins numerous high-level vision applications. Recent advances in Transformer architectures have significantly improved edge quality by capturing long-range dependencies, but this often comes with computational overhead. Achieving higher pixel-level accuracy requires increased input resolution, further escalating computational cost and limiting practical deployment. Building on the strong representational capacity of recent Transformer-based edge detectors, we propose an \textbf{A}daptive \textbf{M}ulti-stage non-edge \textbf{P}runing framework for \textbf{E}dge \textbf{D}etection (\textbf{Amped}). Amped identifies high-confidence non-edge tokens and removes them as early as possible to substantially reduce computation, thus retaining high accuracy while cutting GFLOPs and accelerating inference with minimal performance loss. Moreover, to mitigate the structural complexity of existing edge detection networks and facilitate their integration into real-world systems, we introduce a simple yet high-performance Transformer-based model, termed \textbf{S}treamline \textbf{E}dge \textbf{D}etector (\textbf{SED}). Applied to both existing detectors and our SED, the proposed pruning strategy provides a favorable balance between accuracy and efficiency—reducing GFLOPs by up to 40\% with only a 0.4\% drop in ODS F-measure. In addition, despite its simplicity, SED achieves a state-of-the-art ODS F-measure of 86.5\%. The code will be released.
\end{abstract}

\section{Introduction}\label{sec:intro}
Edge detection is a foundational task in computer vision. By locating discontinuities in brightness, texture, or color, it delineates object boundaries and other salient structures~\cite{liu2019richer,xie2015holistically}, providing cues for downstream tasks such as image segmentation~\cite{cheng2016hfs,yu2018learning,li2020improving}, object detection~\cite{ullman1991recognition,ferrari2008groups}, generic object proposal generation~\cite{uijlings2013selective,zitnick2014edge,zhang2017sequential}, and weakly supervised learning~\cite{wei2016stc,hou2017bottom,liu2020leveraging}. Thus, improving the performance of edge detection models remains critical to advancing the overall understanding and analytical power of visual systems.

\begin{figure}[!t]
    \centering
    \includegraphics[width=.5\textwidth]{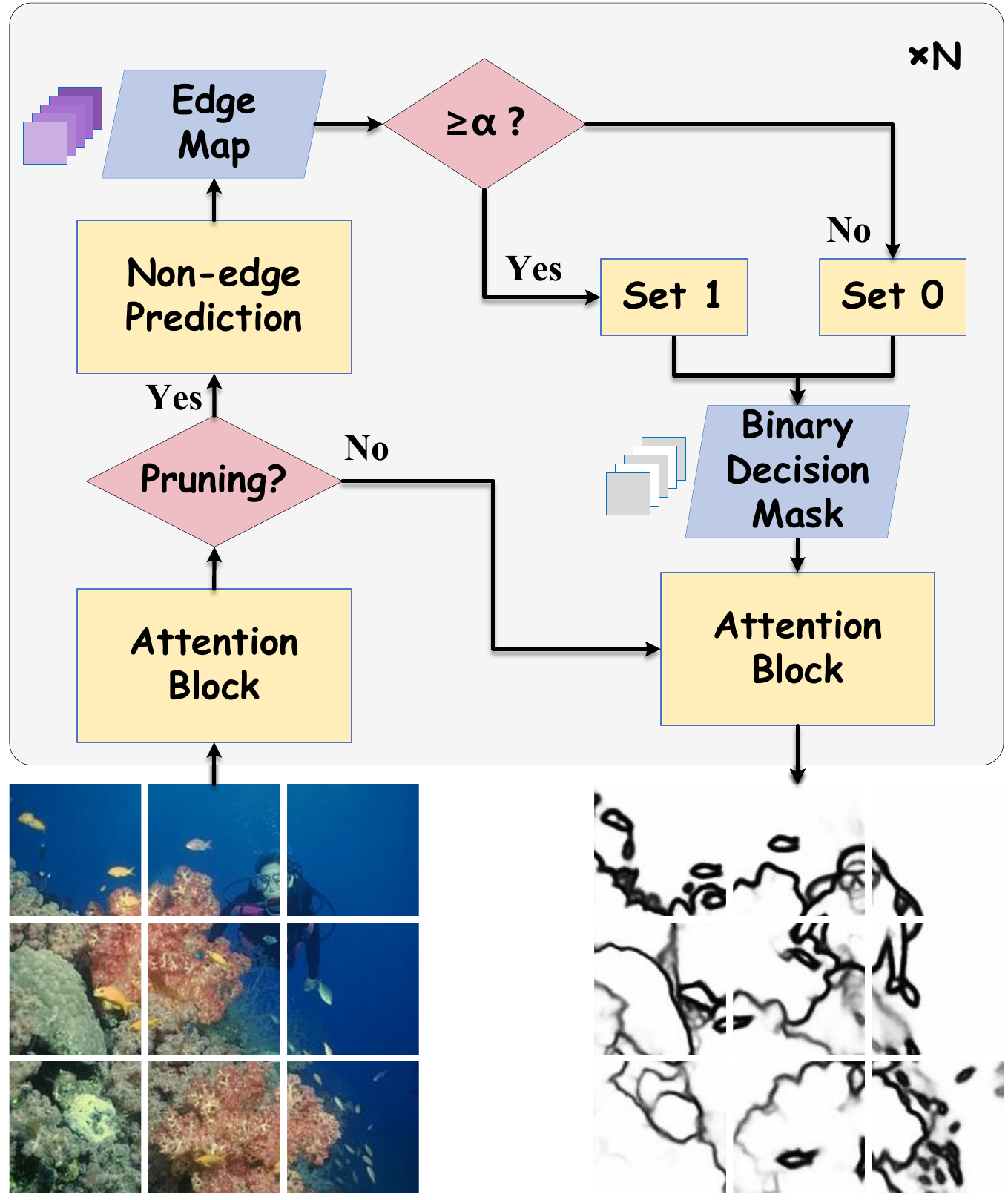}
    \caption{\textbf{Non-edge pruning flowchart.} As the backbone network extracts features, our method adaptively generates a binary decision mask by thresholding the intermediate edge score maps to determine which tokens should be pruned.}
    \label{fig:flowchart}
    \vspace{-.2in}
\end{figure}

Since the introduction of the Vision Transformer (ViT)~\cite{dosovitskiy2021image}, transformer-based edge detection methods \cite{pu2022edter,ma2023eatder} have demonstrated clear advantages, owing to their strong capability for global context modeling and capturing long-range dependencies \cite{sun2024rethinking}. 
They yield significantly sharper and more consistent edge maps, especially in scenes with complex structures, diverse textures, or heavy occlusion. Recent methods achieve state-of-the-art results on BSDS500 \cite{arbelaez2010contour} and NYUDv2 \cite{silberman2012indoor} datasets, outperforming conventional convolutional neural network (CNN)-based methods \cite{xie2015holistically,liu2019richer,he2019bi} by a large margin in both accuracy and robustness.

While transformer-based edge detection methods have manifested significant advantages in edge accuracy, the substantial computational cost and high latency introduced when processing high-resolution images severely hinder their practical deployment on resource-constrained devices, such as mobile or embedded systems. Therefore, to enhance model efficiency while preserving high accuracy, it is crucial to lightweight these models, which is typically achieved through model compression techniques and token pruning is notably effective among them. However, most token pruning methods are designed for image classification \cite{rao2021dynamicvit,xu2022evo,yin2022vit}, which is fundamentally different from edge detection. Image classification is an image-level prediction robust to the loss of individual tokens, whereas edge detection is a dense per-pixel task requiring rich spatial details. Existing indiscriminate token pruning can thus impair the model's ability to capture fine edges. The core technical difficulty, therefore, lies in determining \textit{how to accurately evaluate the importance of each token for pruning.}

Furthermore, the inherent multi-scale feature utilization in edge detection adds another layer of complexity to pruning. An efficient model leverages features from different transformer layers, each with a distinct role: shallow layers capture fine-grained, low-level intensity changes, while deep layers integrate high-level semantics for object-boundary recognition. Both are crucial for generating complete and accurate edge maps. This characteristic demands a more nuanced pruning strategy, presenting a core challenge: \textit{how to design a pruning criterion that adapts to the distinct contributions of features at different hierarchies?}

Finally, we observe that existing transformer-based edge detectors often rely on structurally complex decoders, \eg, elaborately designed CNNs for feature fusion and upsampling. These intricate architectures not only introduce substantial computational overhead but also pose two significant practical obstacles: the complex internal information flow impedes the effective application of importance score-based adaptive pruning, and the overall design considerably increases the difficulty of integrating and deploying the model in real-world, complex systems. This raises the final technical challenge: \textit{how to design a simple yet effective edge detector that facilitates efficient pruning?}

To address the aforementioned technical challenges, we propose an \textbf{A}daptive \textbf{M}ulti-stage non-edge \textbf{P}runing framework for \textbf{E}dge \textbf{D}etection (\textbf{Amped}), as shown in \cref{fig:flowchart}. The key insight behind Amped is that the vast majority of image pixels correspond to non-edge regions, implying that high-confidence non-edge tokens can be safely pruned with minimal risk of error. Accordingly, we compute an edge score map to estimate token importance, and then apply thresholding to obtain a binary decision mask that preserves critical edge tokens while pruning redundant ones. This mask dynamically guides the attention computation in subsequent transformer layers, thereby enabling efficient, data-dependent pruning. To tackle the hierarchical nature of edge detectors, Amped exploits intrinsic edge score predictions from different layers as hierarchical supervision to progressively refine pruning decisions. Specifically, at each stage, a preliminary edge map is generated to produce a binary decision mask that identifies pruning candidates for the next stage, forming an adaptive multi-stage pruning process. Finally, to improve practical deployment feasibility, we design a \textbf{S}treamline \textbf{E}dge \textbf{D}etector (\textbf{SED}) architecture that that achieves high accuracy with reduced computational load, making it inherently compatible with model pruning.

Our contributions can be summarized as follows:
\begin{itemize}
    \item We introduce a novel non-edge pruning strategy that explicitly identifies and removes redundant non-edge tokens while preserving critical tokens for edge prediction, enabling fine-grained pruning tailored to edge detection.
    \item We develop an adaptive multi-stage non-edge pruning method to progressively refine pruning decisions across transformer layers, effectively aligning with the hierarchical nature of edge detection.
    \item We propose a streamline yet effective transformer-based model, SED, which surpasses existing edge detectors by a significant margin and is compatible with model pruning, thereby benefiting practical deployment.
\end{itemize}

\section{Related Work}
\subsection{Edge Detection}
Edge detection aims to identify object boundaries and salient contours, with the key challenge being to balance localization accuracy, robustness to noise, and contour continuity. Early methods~\cite{roberts1963machine,sobel1972camera,canny1986computational} primarily rely on hand-crafted differential operators, most notably the Canny detector~\cite{canny1986computational}. With the advent of machine learning, statistical methods~\cite{martin2004learning,dollar2006supervised,arbelaez2010contour,ren2012discriminatively,dollar2014fast,hallman2015oriented} such as gPb~\cite{arbelaez2010contour} improve accuracy by integrating multiple local features with classifier-based strategies.

Deep learning has transformed this field~\cite{bertasius2015deepedge,shen2015deepcontour,bertasius2015high,maninis2017convolutional,wang2017deep,he2019bi,zhou2024muge}. Fully convolutional networks, such as HED~\cite{xie2015holistically} and RCF~\cite{liu2019richer}, enable end-to-end edge learning and aggregate multi-scale features through hierarchical fusion. Building on this success, Transformer models have emerged as strong backbones. ViT \cite{dosovitskiy2021image} uses self-attention to model global context and mitigates CNNs' limitations in capturing long-range dependencies. Consequently, numerous Transformer designs~\cite{pu2022edter,ye2024diffusionedge,jie2024edgenat} combine CNN's local feature extraction with Transformer's global context modeling to improve edge detection accuracy. However, the self-attention mechanism incurs quadratic computational complexity with respect to the input length. To this end, this work exploits model pruning to improve the efficiency of Transformer-based edge detectors.

\subsection{Model Compression}
Model compression has evolved from static, coarse-grained strategies to dynamic, fine-grained approaches. Early works emphasize post-training methods such as weight pruning~\cite{han2015learning,luo2017thinet}, quantization~\cite{jacob2018quantization}, and low-rank decomposition~\cite{kim2016compression,denton2014exploiting}. These techniques reduce model size but often deliver limited compression at fixed accuracy, offer restricted flexibility, and exhibit poor hardware affinity. As deep learning rises, structured pruning~\cite{he2017channel,liu2017learning,xia2022structured} make a shift toward systematic, hardware-aware compression and acceleration.

With the widespread adoption of the Transformer \cite{vaswani2017attention,dosovitskiy2021image}, research has shifted to optimizing attention and developing dynamic compression for input-adaptive, fine-grained computation~\cite{fan2020reducing}, such as token pruning~\cite{goyal2020power} and early exiting~\cite{xin2020deebert}. Notably, token pruning removes redundant tokens in Transformer while retaining informative ones for subsequent computation, and has achieved considerable success in recent studies. Against this backdrop, recent work focuses on improving efficiency without sacrificing accuracy to meet the low-latency, low-power, and high-throughput constraints of mobile edge devices and real-time vision applications. Existing token pruning research, however, has largely concentrated on image classification, where substantial speedups and compression have been reported \cite{rao2021dynamicvit,yin2022vit,xu2022evo}. In this work, we aim to bridge this gap by introducing token pruning specifically tailored for edge detection models, improving efficiency while preserving accuracy.

\begin{figure*}[!t]
    \centering
    \includegraphics[width=\textwidth]{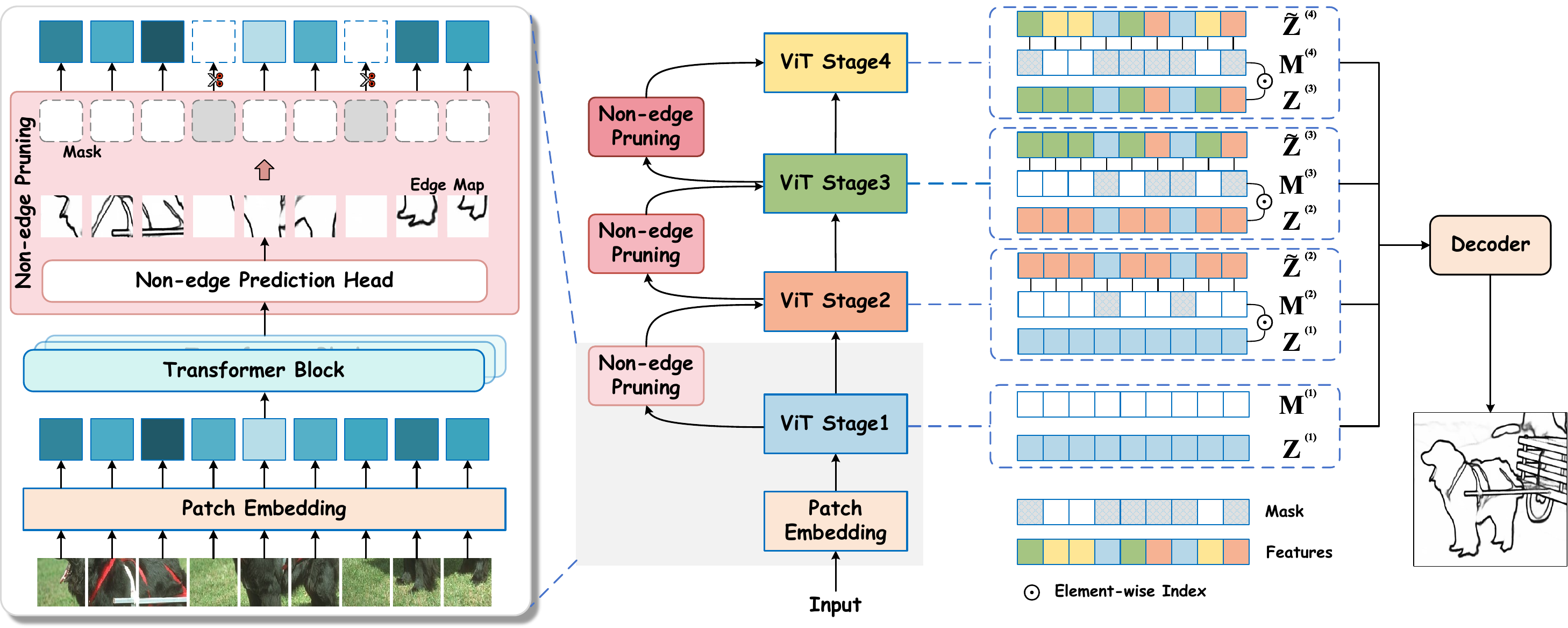}
    \caption{\textbf{Overview of the proposed Amped framework}. Amped trims high-confidence non-edge tokens by computing the edge score map and comparing it against a stage-specific threshold. The proposed SED utilizes a simple linear decoder, a design that effectively reduces model complexity while maintaining high precision. $\mathbf{Z}^{(l)}$ and $\mathbf{M}^{(l)}$ denote the feature map and binary decision mask of pruning stage $l$, and $\tilde{\mathbf{Z}}^{(l)}$ represents the recovered feature map of stage $l$.}
    \label{fig:framework}
\end{figure*}

\section{Method}
\cref{fig:framework} illustrates our Amped framework. In \cref{subsec: non-edge_pruning}, we detail our non-edge pruning method, which generates edge score maps at intermediate layers to guide the pruning of high-confidence non-edge tokens. \cref{subsec: multi-stage_pruning} presents our adaptive multi-stage pruning strategy, which dynamically adjusts pruning intensity across different stages according to the distinct information focus of shallow versus deep features. Finally, \cref{sec:SED} introduces our structurally simple SED model, which attains high-precision edge detection.

\subsection{Non-edge Pruning}
\label{subsec: non-edge_pruning}
Edge detection can be regarded as a specialized form of pixel-wise dense classification. The output is divided into edge pixels rich in structural information and non-edge pixels lacking critical contour details. However, unlike ViT-based image classification \cite{dosovitskiy2021image,liu2024vision}, which is empirically robust to aggressive pruning of low-importance tokens~\cite{rao2021dynamicvit,liang2022not}, edge detection is highly sensitive to spatial details. Thus, incorrect pruning of tokens directly leads to severe issues such as fragmented contours and blurry boundaries, significantly degrading output quality. Striking the crucial balance between preserving critical edges and reducing computational load, our pruning method specifically targets the numerically dominant yet unevenly distributed non-edge tokens, achieving significant computational efficiency gains while maintaining fine-grained detail preservation.

To separate non-edge tokens from edge tokens, we employ a lightweight non-edge prediction head at each pruning stage. For a given pruning stage $l$ with input $\mathbf{X}^{(l)} \in \mathbb{R}^{N_l \times C}$, which has $N_l$ tokens and each token has $C$ channels, the head is a linear layer with weight $\mathbf{W}_{\mathrm{pred}}^{(l)}$ followed by a sigmoid function $\sigma(\cdot)$, which produces a token-wise edge score vector:
\begin{equation}
\label{eq:nepm_prob}
    \mathbf{P}^{(l)} = \sigma \left(\mathbf{X}^{(l)}\mathbf{W}_{\mathrm{pred}}^{(l)} \right ),
\end{equation}
where $\mathbf{P}^{(l)}$ stores the edge score of each token $p_i^{(l)}$.

By thresholding the token-wise scores $\mathbf{P}^{(l)}$ with a stage-specific threshold $\alpha_l$ as constrained in \cref{eq:relationship}, we obtain a binary decision mask $\mathbf{M}^{(l)} \in \mathbb{R}^{N_l \times 1}$:
\begin{equation}
\label{eq:stage_mask}
\mathbf{M}^{(l)}_{i} =
\begin{cases}
1, & \text{if } p_i^{(l)} \ge \alpha_l,\\
0, & \mathrm{otherwise},
\end{cases}
\quad \mathbf{M}^{(l)}_{i} \in \{0,1\},
\end{equation}
where $\mathbf{M}^{(l)}_i = 1$ indicates a retained token and $\mathbf{M}^{(l)}_i = 0$ indicates a pruned token. 
For high-precision dense prediction tasks, during subsequent attention computation in Transformer, pseudo-inhibition can cause numerical precision to leave small residuals that perturb the self-attention calculation. To avoid this effect, in models like ViT \cite{dosovitskiy2021image}, we use hard pruning strategy and explicitly remove pruned tokens from the attention computation. Let $\mathbf{I}^{(l)}_{M} = \{ i \in [1,N_l]\mid \mathbf{M}^{(l)}_i = 1 \}$ denote the index set of tokens selected by the binary mask $\mathbf{M}^{(l)}$, $N_l' = |\mathbf{I}^{(l)}|$ be the length of remaining tokens and $\pi_M:\mathbb{R}^{N_{l} \times C}\rightarrow \mathbb{R}^{N_{l}'\times C}$ perform as a projection operator defined by the mask $\mathbf{M}^{(l)}$.
\begin{equation}
\tilde{\mathbf{X}}^{(l)} = \pi_M(\mathbf{X}^{(l)}) = \left\{ x_i^{(l)} \mid i \in \mathbf{I}_M^{(l)} \right\}
\end{equation}
We then proceeds to the multi-head self-attention computation, with $\tilde{\mathbf{Q}}^{(l)}$, $\tilde{\mathbf{K}}^{(l)}$ and $\tilde{\mathbf{V}}^{(l)}$ standing for matrices computed from trimmed $\tilde{\mathbf{X}}^{(l)}$, and $\mathbf{Z}^{(l)}$ denoting the output feature map of pruning stage $l$:
\begin{equation}
\label{eq:attn_logits_full}
    \mathbf{A}^{(l)} = \frac{\tilde{\mathbf{Q}}^{(l)} {\tilde{\mathbf{K}}^{(l)\,\top}}}{\sqrt{C}} \in \mathbb{R}^{N_l' \times N_l'},
\end{equation}
\begin{equation}
    \mathbf{Z}^{(l)}_{i} = \sum_{j=1}^{N_{l}'} \frac{\exp(\mathbf{A}^{(l)}_{i,j})}{\sum_{k=1}^{N_{l}'}\exp(\mathbf{A}^{(l)}_{i,k})}\tilde{\mathbf{V}}^{(l)}_j.
\end{equation}
$\mathbf{Z}^{(l)}$ is fed into following layers for calculation until the next pruning operation. 

Upon completing the backbone processing, we proceed to the decoder, which operates on a full-length token sequence to predict the edge probability map. To this end, the sequence $\mathbf{Z}^{(l)}$ must be restored to its original length. We achieve this by accumulating the pruning masks across network depth. Specifically, let $\tilde{\mathbf{M}}^{(l)} \in \mathbb{R}^{N \times 1}$ denote the pruning mask with respect to the original full-length tokens at the $l$-th pruning stage, in which $N$ is the number of tokens before pruning. The mask is updated using $\mathbf{M}^{(l)}$ as follows:
\begin{equation}
    \label{eq:mask_accum}
    \tilde{\mathbf{M}}^{(l)} = \tilde{\mathbf{M}}^{(l-1)}, \qquad\quad
    \tilde{\mathbf{M}}^{(l)}_{i_k} = \mathbf{M}^{(l)}_k, \quad
    \wrt~ i_k \in \{i | \tilde{\mathbf{M}}^{(l)}_i = 1 \},
\end{equation}
where $\tilde{\mathbf{M}}^{(0)} = \mathbf{1}_N$ is defined as an all-ones vector. Then, we scatter the retained tokens back to their original positions according to $\tilde{\mathbf{M}}^{(l)}$. Let $\tilde{\mathbf{Z}}^{(l)}$ represent the recovered full-length token sequence for stage $l$. Starting from stage $l$, we set $\tilde{\mathbf{Z}}^{(l)} = \mathbf{Z}^{(l)}$ and the reconstruction is performed as:
\begin{equation}
    \label{eq:recovery_scatter}
    \tilde{\mathbf{F}}^{(l)} = \mathbf{X}^{(t)}, 
    \qquad
    \tilde{\mathbf{F}}^{(l)}_{i_k} = \tilde{\mathbf{Z}}^{(l)}_k, ~
    \wrt~ i_k \in \{i | \tilde{\mathbf{M}}^{(t)}_i = 1 \},\qquad
    \tilde{\mathbf{Z}}^{(l)} = \tilde{\mathbf{F}}^{(l)},
\end{equation}
which is iterated for $t=l, l-1, \cdots, 2, 1$ to get final $\tilde{\mathbf{Z}}^{(l)} \in \mathbb{R}^{N \times C}$. Repeating this procedure for all stages $l$ yields full-length token sequences for the decoder.

\subsection{Adaptive Multi-Stage Non-edge Pruning}
\label{subsec: multi-stage_pruning}
Transformer-based edge detection models rely on multi-scale features with distinct characteristics~\cite{liu2019richer,pu2022edter}: shallow features capture precise localization despite semantic ambiguity, whereas deep features provide clear semantics at the cost of coarse localization. This inherent property necessitates effective cross-level integration, so we introduce an adaptive multi-stage pruning strategy that dynamically evaluates token importance at different transformer stages according to their intrinsic characteristics, particularly their predicted edge maps. 

Specifically, in the shallow layers of the network, spatial details in the feature maps are crucial for precise edge localization, even though the predicted edge maps at this stage are often noisy and of low confidence. Using a high threshold here would prematurely discard valuable localization cues. We therefore adopt a lower threshold in shallow layers, implementing a more conservative pruning strategy to preserve spatial details. In contrast, deep layers encode features rich in high-level semantic information, whose primary role is semantic contour recognition rather than pixel-level localization, and their edge predictions are typically clearer and of higher confidence. Consequently, we set a higher threshold in deep layers, enabling more extensive yet reliable pruning of tokens confidently predicted as non-edge, thereby improving computational efficiency.

Therefore, for a network with $L$ pruning stages, stage-specific thresholds $\{\alpha_l\}_{l=1}^{L}$ is aimed to approximate a non-decreasing sequence along the depth as closely as possible, which is mathematically expressed as:
\begin{equation}
    \label{eq:relationship}
    \alpha_{1}\leq \alpha_{2} \leq \dots \leq \alpha_{L}, ~\alpha_{l} \in [0,1].
\end{equation}
This approach conservatively preserves spatial details in shallow layers while aggressively eliminating semantic redundancy in deep layers, ultimately achieving an optimal balance between accuracy and computational efficiency.

\subsection{Streamline Edge Detector}
\label{sec:SED}
Current edge detection models predominantly employ a Transformer encoder coupled with a heavy, computationally intensive decoder, which hinders effective compression and deployment and complicates integration into broader vision systems due to its rigid design. To address these limitations, we design a Streamline Edge Detection (SED) model that achieves superior accuracy while substantially reducing architectural complexity. This SED architecture yields a dual advantage of deployment efficiency and modular flexibility, supporting straightforward integration into general-purpose vision pipelines and thus fostering the practical adoption of edge detection in real-world scenarios.

We adapt the SAM-ViT~\cite{kirillov2023segment} backbone for SED to build \textbf{SED-SViT}, leveraging its strong low-level feature representations while circumventing its semantic bottlenecks. Unlike conventional edge detectors trained on closed datasets with limited scene diversity, SAM is pre-trained at scale via a supervised-to-automatic mask annotation pipeline without predefined categories, yielding robust open-world boundary representations \cite{ariff2026evaluating}. However, SAM is designed for prompt mask prediction rather than rich semantic reasoning, which induces a strong representational bias toward generic contours and region boundaries and makes it difficult to repurpose the model for heterogeneous downstream tasks. In practice, existing follow-up works usually use SAM as a frozen or lightly adapted feature extractor or mask generator~\cite{huang2024alignsam,wu2025medical}, and the few fine-tuning efforts largely remain within generic segmentation variants~\cite{gu2025build,peng2024sam} rather than extending to substantially different downstream tasks. Motivated by this, we deliberately discard the complex prompt-conditioned mask decoder and retain only the SAM-ViT image encoder as our backbone, turning SAM into a boundary-sensitive yet streamline architecture, SED-SViT, which is amenable to our pruning framework. In our experiments, we additionally construct \textbf{SED-ViT} by replacing SAM-ViT~\cite{kirillov2023segment} in SED-SViT with vanilla ViT \cite{dosovitskiy2021image} for comparison with SED-SViT.

During decoding, we deliberately avoid heavy decoders and adopt a lightweight multi-scale fusion head that operates directly on the recovered full-length token sequences. Let \(\tilde{\mathbf{Z}}^{(l)}\) represent the recovered full-length token sequence at stage \(l\). Each stage is first passed through a stage-specific linear projection:
\begin{equation}
    \label{eq:dec_stage_linear}
    \hat{\mathbf{Z}}^{(l)} = f_{\mathrm{dec}}^{(l)}\!\left(\tilde{\mathbf{Z}}^{(l)}\right),
    \quad l = 0,\dots,L,
\end{equation}
where \(f_{\mathrm{dec}}^{(l)}(\cdot)\) is a learnable linear layer used to align channel information across different stages. We then fuse multi-scale information by concatenating the projected features from all stages along the channel dimension:
\begin{equation}
    \label{eq:dec_fuse}
    \mathbf{Z}_{\mathrm{fuse}} = 
    \left[\, \hat{\mathbf{Z}}^{(1)};\, \hat{\mathbf{Z}}^{(2)};\, \dots;\, \hat{\mathbf{Z}}^{(L)} \right].
\end{equation}
Finally, we apply a final projection followed by normalization and point-wise activation to obtain the final output tokens used for edge map prediction:
\begin{equation}
    \label{eq:dec_out}
    \mathbf{Z}_{\mathrm{out}} 
    = \phi\!\Big(
        \mathrm{Norm}\big(\mathbf{Z}_{\mathrm{fuse}} \mathbf{W}_{\mathrm{proj}} + \mathbf{b}_{\mathrm{proj}}\big)
      \Big),
\end{equation}
where \(\mathbf{W}_{\mathrm{proj}}\) and \(\mathbf{b}_{\mathrm{proj}}\) are learnable projection parameters and \(\phi(\cdot)\) implies a point-wise activation.

\section{Experiments}
\subsection{Experimental Setup}
\subsubsection{Datasets.}
We evaluate the proposed method on two well-known edge detection benchmarks: BSDS500~\cite{arbelaez2010contour} and NYUDv2~\cite{silberman2012indoor}.
\textbf{BSDS500}~\cite{arbelaez2010contour} contains 500 natural images split into 200 for training, 200 for validation, and 100 for testing. Each image is annotated by multiple annotators, yielding rich and highly consistent boundary ground truth. Following common practice \cite{liu2019richer,he2019bi,pu2022edter,ye2024diffusionedge}, we pretrain the model on the PASCAL VOC Context dataset \cite{mottaghi2014role}, which provides 10,103 images with pixel-level annotations covering complete scenes and high-quality semantic boundaries. The dataset's diverse scenes make it a strong pretraining source for edge detection.
\textbf{NYUDv2}~\cite{silberman2012indoor} is an indoor RGB-D dataset containing 1{,}449 images, \ie, 795 for training and 654 for testing. It provides dense pixel-level annotations across diverse indoor scenes. In addition to RGB images, it offers silhouettes derived from depth and surface normals, supplying complementary geometric and appearance cues.

\subsubsection{Implementation Details.}
To comprehensively evaluate our pruning method, we conduct extensive experiments on the state-of-the-art method EDTER~\cite{pu2022edter} and our SED architectures. We use \textbf{SED-SViT-B/L} and \textbf{SED-ViT-B/L} to denote SED-SViT and SED-ViT with base/large SAM-ViT and ViT backbones, respectively. \textbf{SED-SViT-BP/LP} and \textbf{SED-ViT-BP/LP} indicates their corresponding pruned versions.
All experiments are implemented in PyTorch based on the \texttt{MMSegmentation} framework~\cite{mmseg2020}. 
SED-SViT and SED-ViT are initialized from the pretrained weights of SAM-ViT~\cite{kirillov2023segment} and ViT~\cite{dosovitskiy2021image}, respectively. 
For SED-SViT and SED-ViT, the input is resized to $1024\times 1024$ with a large patch size of 16. The batch size is 8 on BSDS500 and 4 on NYUDv2, and all models are trained for 40k iterations on both datasets. All experiments are executed on a computing cluster equipped with two NVIDIA GeForce RTX 4090 GPUs.

\subsubsection{Evaluation Metrics.}
For evaluation, we adopt three standard metrics~\cite{arbelaez2010contour}: ODS (Optimal Dataset Scale) F-measure, OIS (Optimal Image Scale) F-measure, and Average Precision (AP). ODS is computed using a single global threshold selected to maximize the F-measure over the entire test set, whereas OIS selects an optimal threshold for each individual image.
Before scoring, all predicted edge maps are processed with non-maximum suppression \cite{dollar2014fast}. Following common practice~\cite{liu2019richer,he2019bi,pu2022edter,ye2024diffusionedge}, the localization tolerance is set to 0.0075 for the BSDS500 datset~\cite{arbelaez2010contour} and 0.011 for the NYUDv2 dataset~\cite{silberman2012indoor}.

\subsection{Ablation Study}
\subsubsection{Effect of Non-edge Pruning Score.}
We first conduct experiments to determine the optimal pruning score threshold for each model based on its performance. The corresponding quantitative results are presented in \cref{tab:pruning_score}. We aim to achieve favorable trade-off between accuracy and efficiency. Experimental results indicate that the SED-SViT model demonstrates an excellent balance when using pruning scores of $[0.3, 0.4, 0.5]$, while GFLOPs are substantially reduced by 28.07\%, ODS F-measure and OIS F-measure decrease by only 0.6\% and 0.5\%, respectively. For the SED-ViT architecture, a well-proportioned result is obtained with pruning scores of $[0.3, 0.4, 0.5]$, achieving a 38.80\% reduction on GFLOPs with only a 0.8\% and 1.0\% degradation in ODS and OIS, respectively. The data in the table further reveals a consistent pattern: within the bounds of reasonable threshold settings, as pruning score increases, more non-edge tokens are removed, leading to a more significant reduction in GFLOPs. Meanwhile, this aggressive pruning also results in a corresponding decline in accuracy.

\begin{table}[!t]
    \centering
    \caption{Ablation study of non-edge pruning scores for SED-SViT and SED-ViT on BSDS500~\cite{arbelaez2010contour}. All results are obtained using models with Base-scale backbones.}
    \label{tab:pruning_score}
    \resizebox{\columnwidth}{!}{%
    \begin{tabular}{c|c|c|c|c|c|c}
      \toprule
      \multirow{2}{*}{Pruning Scores} & \multicolumn{2}{c|}{SED-SViT} & \multirow{2}{*}{GFLOPs} & \multicolumn{2}{c|}{SED-ViT} & \multirow{2}{*}{GFLOPs} \\
      \cmidrule{2-3}\cmidrule{5-6}
       & ODS & OIS & & ODS & OIS &  \\
      \midrule
      origin & 0.849 & 0.868 & 484.5 & 0.826 & 0.847 & 663.7 \\ \relax
      [0.25, 0.35, 0.45] & 0.843(-0.6) & 0.864(-0.4) & 355.4($\downarrow$26.65\%) & 0.820(-0.6) & 0.838(-0.9) & 407.3($\downarrow$38.63\%)\\ \relax
      [0.3, 0.4, 0.5] & 0.843(-0.6) & 0.863(-0.5) & 348.5($\downarrow$28.07\%)&0.818(-0.8) & 0.837(-1.0) & 406.2($\downarrow$38.80\%) \\ \relax
      [0.35, 0.45, 0.55] & 0.840(-0.9) & 0.860(-0.8) & 339.6($\downarrow$29.91\%) & 0.816(-1.0)  & 0.837(-1.0) & 390.2($\downarrow$41.21\%) \\ \relax
      [0.4, 0.5, 0.6] & 0.824(-2.5) & 0.846(-2.2) & 216.6($\downarrow$55.29\%) & 0.812(-1.4) & 0.831(-1.6) & 332.3($\downarrow$49.93\%) \\
      \bottomrule
    \end{tabular}}
\end{table}

\subsubsection{Effect of Multi-stage Pruning.}
We evaluate the proposed adaptive multi-stage non-edge pruning on SED-SViT and SED-ViT, and the results are presented in \cref{tab:multi-stage}. Pruning exclusively at the 3rd layer for the ViT-Base model \cite{dosovitskiy2021image}, reduces computation by only about 15\%. However, accuracy metrics incur appreciable drops for SED-SViT and SED-ViT, with ODS F-measure decreases 0.5\% and 0.2\% separately, and OIS F-measure decreases 0.4\% and 0.8\%, respectively. Pruning at three levels shows the best overall trade-off with 28.07\% and 38.80\% GFLOPs decline, and 0.6\% and 0.8\% ODS F-measure decline, for SED-SViT and SED-ViT, respectively. These results suggest that shallow features are crucial for edge localization and should be pruned conservatively, whereas deeper layers contain redundant non-edge tokens that can be removed more aggressively, which confirms the effectiveness of our adaptive multi-stage pruning.

\begin{table}[!t]
    \centering
    \caption{Ablation study of multi-stage pruning for SED-SViT and SED-ViT on BSDS500~\cite{arbelaez2010contour}. All results are obtained using models with Base-scale backbones and pruning scores of [0.3, 0.4, 0.5].}
    \label{tab:multi-stage}
    \resizebox{\columnwidth}{!}{%
    \begin{tabular}{c|c|c|c|c|c|c}
      \toprule
      \multirow{2}{*}{Pruning Layers} & \multicolumn{2}{c|}{SED-SViT} & \multirow{2}{*}{GFLOPs} & \multicolumn{2}{c|}{SED-ViT} & \multirow{2}{*}{GFLOPs} \\
      \cmidrule{2-3}\cmidrule{5-6}
       & ODS & OIS & & ODS & OIS &  \\
      \midrule
      origin & 0.849 & 0.868 & 484.5 & 0.826 & 0.847 & 663.7 \\
      3 & 0.844(-0.5) & 0.864(-0.4) & 411.7($\downarrow$15.03\%)&0.820(-0.6)& 0.839(-0.8) &566.2($\downarrow$14.69\%)\\
      3, 6 & 0.841(-0.8) & 0.861(-0.7) & 356.0($\downarrow$26.52\%) &0.816(-1.0)  & 0.836(-1.1) &446.9($\downarrow$32.67\%)\\
      3, 6, 9 & 0.843(-0.6) & 0.863(-0.5) & 348.5($\downarrow$28.07\%) & 0.818(-0.8) & 0.837(-1.0) & 406.2($\downarrow$38.80\%)\\
      \bottomrule
    \end{tabular}}
\end{table}

\subsubsection{Effect of Backbone Scale.}
To assess the generalization ability of our proposed non-edge pruning method across backbone scales, we conduct systematic experiments on backbones of varying sizes. The experimental results are presented in \cref{tab:non-edge results}. Compared to the base model, the large model achieves higher accuracy. After pruning, the ODS F-measure improves by 2.0\% and 1.5\% for SED-SViT and SED-ViT, respectively. Meanwhile, the computational cost reduction reflected in GFLOPs changes from 27.97\% to 21.94\% for SED-SViT, and from 38.79\% to 36.41\% for SED-ViT. Our analysis of the results proves the effectiveness and reliability of both our pruning method and the SED model when applied to backbone models of different sizes. The trend in the results reveals a positive correlation between the capacity of the backbone network and its parameter count, with larger models yielding sharper and more coherent edge features. This improvement in accuracy comes at the cost of pruning fewer non-edge tokens, resulting in a more limited reduction in GFLOPs.

\begin{figure}[!t]
    \centering
    \includegraphics[width=\textwidth]{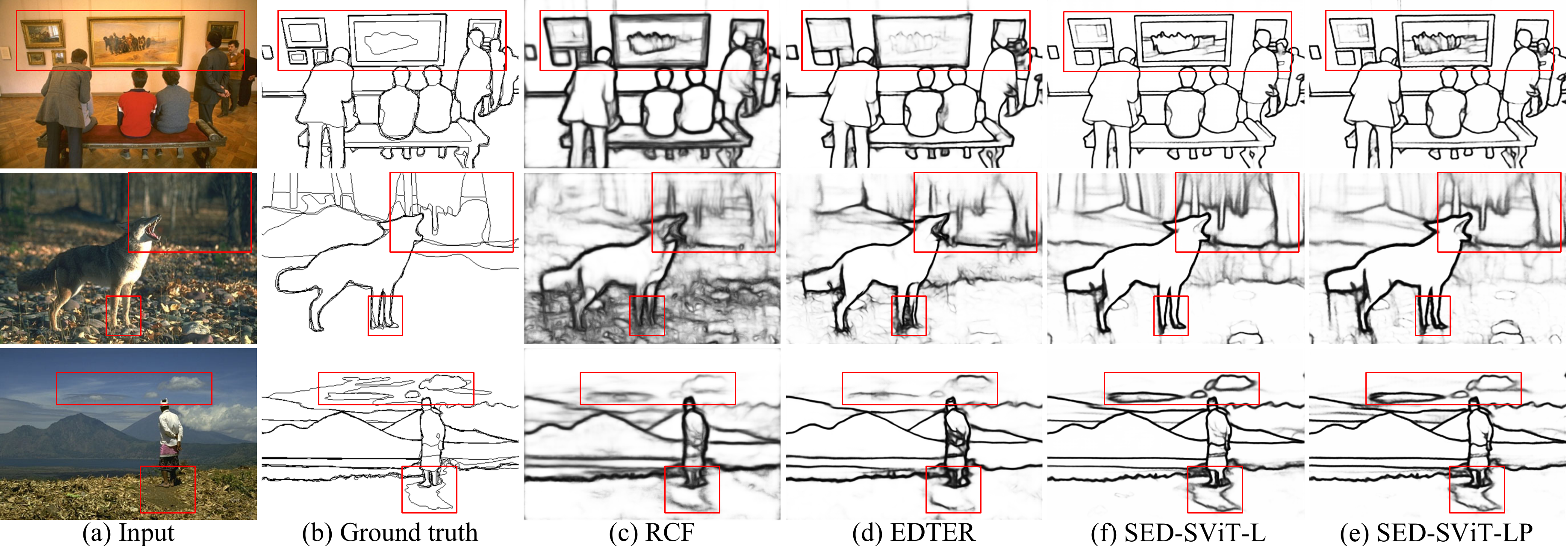}
    \caption{Qualitative comparisons between our method and baselines on three challenging samples from the BSDS500 test set~\cite{arbelaez2010contour}.}
    \label{fig:qualitative_comparisons}
\end{figure}

\begin{table}[!t]
    \centering
    \caption{Results of Amped applied to the prior method EDTER~\cite{pu2022edter} and our models SED-SViT and SED-ViT on BSDS500~\cite{arbelaez2010contour} and NYUDv2~\cite{silberman2012indoor}.}
    \label{tab:non-edge results}
    \resizebox{\columnwidth}{!}{%
    \begin{tabular}{c|c|c|c|c|c|c|c|c|r}
      \toprule
      \multicolumn{2}{c|}{\multirow{2}{*}{Methods}} & \multicolumn{3}{c|}{Origin} &\multirow{2}{*}{GFLOPs}&\multicolumn{3}{c|}{Pruned} & \multicolumn{1}{c}{\multirow{2}{*}{GFLOPs}} \\
      \cmidrule{3-5}\cmidrule{7-9}
      \multicolumn{2}{c|}{} & ODS & OIS & AP & & ODS & OIS & AP &   \\
      \midrule
      \multirow{5}{*}{\rotatebox[origin=c]{90}{BSDS500}} & 
      EDTER & 0.829 & 0.845 & 0.883 & 356.4* & 0.825(-0.4) & 0.839(-0.6) & 0.878(-0.5) & 206.8*($\downarrow$41.98\%) \\
      & SED-ViT-B & 0.826 & 0.847 & 0.884 & 663.7 & 0.818(-0.8) & 0.837(-1.0) & 0.877(-0.7) & 406.2($\downarrow$38.79\%)\\
      & SED-ViT-L & 0.835 & 0.854 & 0.890 & 2069.6 & 0.833(-0.2) & 0.850(-0.4) & 0.888(-0.2) & 1316.1($\downarrow$36.41\%)\\
      & SED-SViT-B & 0.849 & 0.868 & 0.902 & 484.5 & 0.843(-0.6) & 0.863(-0.5) & 0.897(-0.5) & 348.5($\downarrow$28.07\%)\\
      & SED-SViT-L & 0.865 & 0.882 & 0.911 & 1489.2 & 0.863(-0.2) & 0.881(-0.1) & 0.912(+0.1) & 1169.1($\downarrow$21.49\%)\\
      \midrule
      \multirow{5}{*}{\rotatebox[origin=c]{90}{NYUDv2}} & 
      EDTER & 0.758 & 0.771 & 0.747 & 542.5* & 0.752(-0.6) & 0.766(-0.5) & 0.748(+0.1) &  358.4*($\downarrow$33.94\%)\\
      & SED-ViT-B & 0.753 & 0.769 & 0.788 & 663.7 & 0.744(-0.9) & 0.762(-0.7) & 0.776(-1.2) & 416.1($\downarrow$37.31\%)\\
      & SED-ViT-L & 0.764 & 0.781 & 0.799 & 2069.6 & 0.762(-0.2) & 0.780(-0.1) & 0.790(-0.9) & 1342.8($\downarrow$35.12\%)\\
      & SED-SViT-B & 0.779 & 0.797 & 0.806 & 484.5 & 0.772(-0.7) & 0.789(-0.8) & 0.802(-0.4) & 338.0($\downarrow$30.23\%)\\
      & SED-SViT-L & 0.799 & 0.818 & 0.836 & 1489.2 & 0.791(-0.8) & 0.811(-0.7) & 0.819(-1.7) & 1100.1($\downarrow$26.13\%)\\
      \bottomrule
    \end{tabular}}
    \caption*{
        \small
        * For EDTER \cite{pu2022edter}, the reported GFLOPs \textit{exclude} its prediction heads due to their substantial computational cost, \ie, 3503.3 GFLOPs in stage I and 4922.9 GFLOPs in stage II, primarily caused by deconvolution layers with extremely large kernels.
    }
    \vspace{-.3in}
\end{table}

\subsection{Comparison with State-of-the-arts}
\subsubsection{BSDS500.}
We apply the proposed non-edge pruning to both the existing Transformer-based edge detection model EDTER \cite{pu2022edter} and our SED model. Quantitative results are presented in \cref{tab:non-edge results}. On the BSDS500 dataset~\cite{arbelaez2010contour}, pruning the EDTER model leads to a decrease of 0.4\% in ODS F-measure, 0.6\% in OIS F-measure, and 0.5\% in AP, while the GFLOPs of the encoder-decoder network is reduced by 42.17\%. On our SED-ViT method, for the base model, all accuracy metrics drop by less than 1\% with a 38.79\% reduction in GFLOPs, and for the large model, all metrics drop by less than 0.4\% with a 36.41\% reduction in GFLOPs. For SED-SViT, pruning results in accuracy drops of less than 0.6\% for the base model with a 27.97\% GFLOPs reduction, and less than 0.2\% for the large model while still achieving a 21.49\% GFLOPs reduction.

\begin{figure}[!t]
    \centering
    \includegraphics[width=0.8\textwidth]{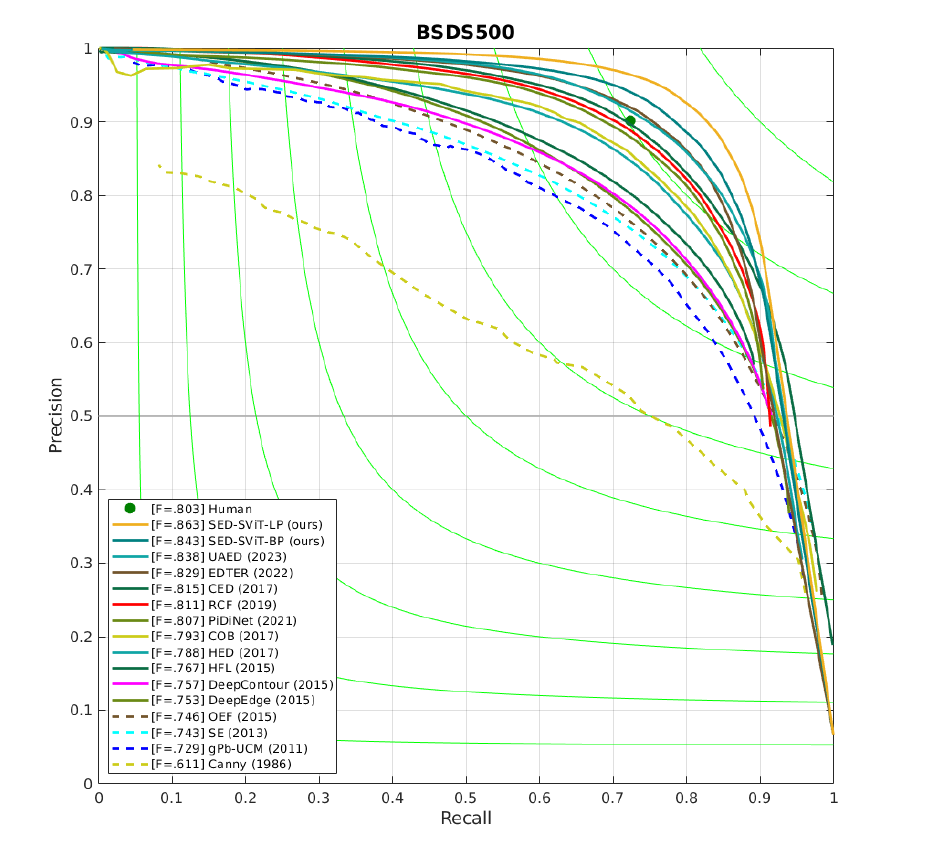}
    \caption{Precision-recall curves on the BSDS500 test set~\cite{arbelaez2010contour}.}
    \label{fig:pr_curve}
\end{figure}

\begin{figure}[!t]
    \centering
    \includegraphics[width=.80\textwidth]{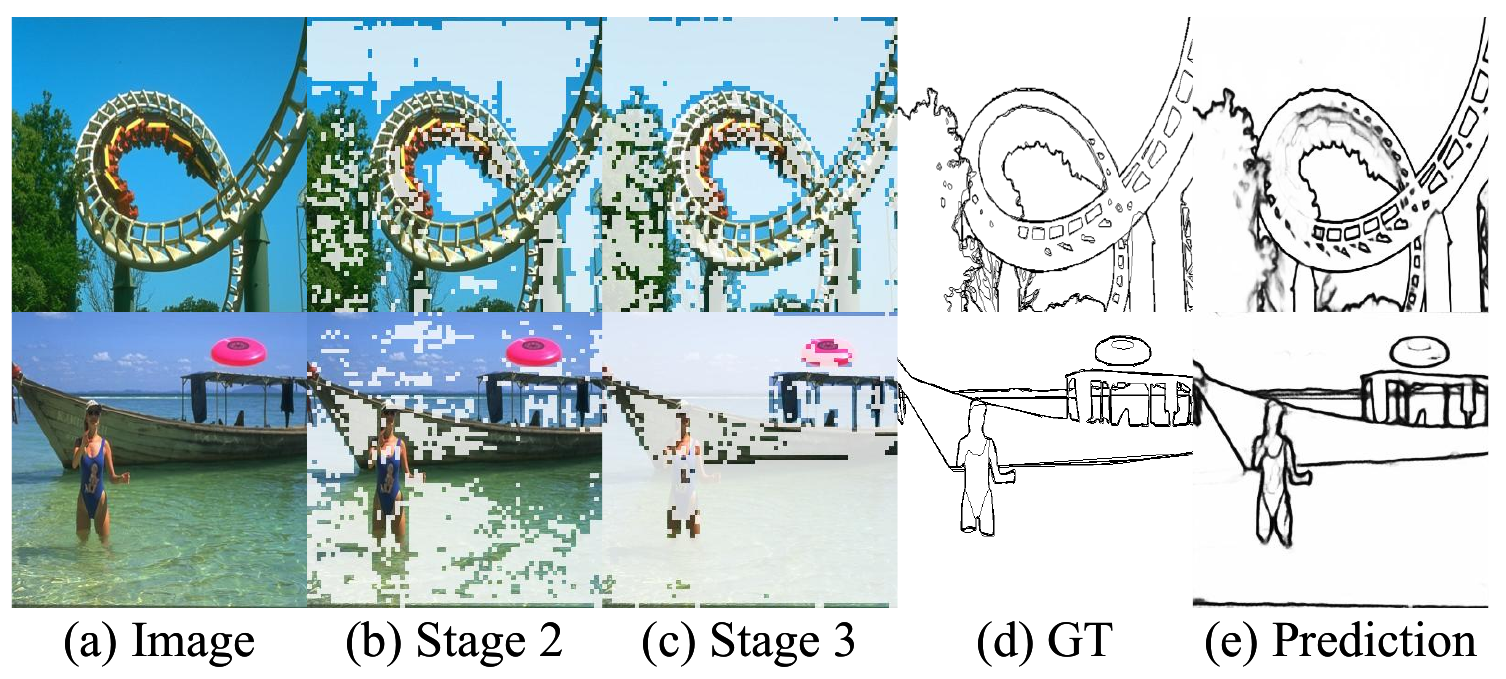}
    \caption{Visualization of progressive non-edge pruning for SED-SViT (top) and SED-ViT (bottom). Most pruned tokens lie in smooth background regions, while tokens near object boundaries are largely preserved, especially for SED-SViT.}
    \label{fig:pruning}
\end{figure}

In \cref{tab:BSDS500 quantitative}, we compare our proposed SED model with previous methods, including DeepEdge~\cite{bertasius2015deepedge}, DeepContour~\cite{shen2015deepcontour}, HFL~\cite{bertasius2015high}, HED~\cite{xie2015holistically}, COB~\cite{maninis2017convolutional}, RCF~\cite{liu2019richer}, CED~\cite{wang2017deep}, BDCN~\cite{he2019bi}, DSCD~\cite{deng2020deep}, PiDiNet~\cite{su2021pixel}, UAED~\cite{zhou2023treasure}, PEdger-L~\cite{fu2023practical}, EDTER~\cite{pu2022edter}, and DiffEdge~\cite{ye2024diffusionedge}. Our approach surpasses the state-of-the-art results with 86.3\% ODS F-measure, 88.1\% OIS F-measure, and 91.2\% AP under single-scale testing. The precision-recall curves of our methods and baselines are plotted in \cref{fig:pr_curve}. In addition, \cref{fig:qualitative_comparisons} shows the qualitative results before and after pruning, comparing to previous models and demonstrating that our method detects clearer, more continuous, and more accurate edges. In \cref{fig:pruning}, we visualize the progressive non-edge pruning process. As can be observed, our method can prune most non-edge regions without intermediate edge signal supervision.

\begin{table}[!t]
    \centering
    \setlength{\tabcolsep}{3.0mm}
    \caption{Quantitative comparison on the BSDS500 test set~\cite{arbelaez2010contour}. All results are evaluated with single scale input.}
    \label{tab:BSDS500 quantitative}
    \resizebox{\columnwidth}{!}{%
    \begin{tabular}{c|l|c|c|c|c}
      \toprule
      Encoder Types & Methods & Pub.'Year & ODS & OIS & AP \\
      \midrule
      \multirow{4}{*}{Traditional}& Canny \cite{canny1986computational} & TPAMI'86 & 0.600 & 0.640 & 0.580\\
      & gPb-UCM \cite{arbelaez2010contour} & TPAMI'10 &0.726 & 0.757 & 0.696\\
      & SE \cite{dollar2014fast} & TPAMI'14 & 0.746 & 0.767 & 0.803\\
      & OEF \cite{hallman2015oriented} & CVPR'15 & 0.746 &0.770 & 0.820 \\
      \midrule
      \multirow{12}{*}{CNN} & DeepEdge \cite{bertasius2015deepedge} & CVPR'15 & 0.753 & 0.772 & 0.807\\
      & DeepContour \cite{shen2015deepcontour} & CVPR'15 & 0.757 & 0.776 & 0.800\\
      & HFL \cite{bertasius2015high} & ICCV'15 & 0.767 & 0.788 & 0.795\\
      & HED \cite{xie2015holistically} & ICCV'15 & 0.788 & 0.808 & 0.840\\
      & COB \cite{maninis2017convolutional} & TPAMI'17 & 0.793 & 0.820 & 0.859\\
      & RCF \cite{liu2019richer} & CVPR'17 & 0.811 & 0.830 & -\\
      & CED \cite{wang2017deep} & CVPR'17 & 0.815 & 0.833 & 0.889\\ 
      & BDCN \cite{he2019bi}& CVPR'19 & 0.828 & 0.844 & 0.890\\
      & DSCD \cite{deng2020deep} & ACMMM'20 & 0.822 & 0.859 & --\\
      & PiDiNet \cite{su2021pixel} & ICCV'21 & 0.807 & 0.823 & -\\
      & UAED \cite{zhou2023treasure} & CVPR'23 & \textbf{\textcolor{blue}{0.838}} & \textbf{\textcolor{blue}{0.855}} & \textbf{\textcolor{blue}{0.902}} \\
      & PEdger-L \cite{fu2023practical} & ACMMM'23 & 0.823 & 0.841 & - \\
      \midrule
      \multirow{4}{*}{Transformer} & EDTER \cite{pu2022edter} & CVPR'22 & 0.829 & 0.845 & 0.883 \\
      & DiffEdge \cite{ye2024diffusionedge} & AAAI'24 & 0.834 & 0.848 & - \\
      & SED-SViT-BP & - & 0.843 & 0.863 & 0.897 \\
      & SED-SViT-LP & - & \textbf{\textcolor{red}{0.863}} &
      \textbf{\textcolor{red}{0.881}} & \textbf{\textcolor{red}{0.912}} \\
      \bottomrule
    \end{tabular}}
\end{table}

\subsubsection{NYUDv2.}
We evaluate the proposed method on RGB images from the NYUDv2 dataset~\cite{silberman2012indoor}, conducting performance analysis and pruning validation. Quantitative results are summarized in \cref{tab:non-edge results}. After applying non-edge pruning to the EDTER model \cite{pu2022edter}, the ODS F-measure reduces 0.6\% and the OIS F-measure declines 0.5\%. Notably, AP improves by 0.1\%, accompanied by a 33.94\% reduction in encoder-decoder GFLOPs. By pruning unimportant non-edge tokens, we analyze that the model can concentrate on edge tokens rich in information, thereby enhancing the output quality. In our SED method, SED-ViT demonstrates a favorable accuracy-efficiency trade-off after pruning: for the base model, ODS/OIS decline by no more than 0.9\% with a 37.31\% GFLOPs reduction; for the large model, ODS/OIS drops are constrained within 0.2\% while achieving a 35.12\% decrease in GFLOPs. For SED-SViT, the base model exhibits less than 0.8\% accuracy degradation with a 30.23\% GFLOPs cut, and the large model limits accuracy loss to within 0.8\% on ODS/OIS while still reducing GFLOPs by 26.13\%. Moreover, we compare the proposed SED with existing advanced approaches, including HED \cite{xie2015holistically}, RCF \cite{liu2019richer}, AMH-Net \cite{xu2017learning}, LPCB \cite{deng2018learning}, BDCN \cite{he2019bi}, PiDiNet \cite{su2021pixel}, PEdger-L \cite{fu2023practical}, EDTER~\cite{pu2022edter}, and DiffEdge~\cite{ye2024diffusionedge}, in \cref{tab:NYUDv2 quantitative}. Under single-scale testing, our method achieves leading performance with 79.1\% ODS F-measure and 81.1\% OIS F-measure.

\begin{table}[!t]
    \centering
    \setlength{\tabcolsep}{2.5mm}
    \caption{Quantitative comparison on the NYUDv2 test set~\cite{silberman2012indoor}. All results are evaluated with single scale input.}
    \label{tab:NYUDv2 quantitative}
    \begin{tabular}{c|l|c|c|c|c}
      \toprule
      Encoder Types & Methods & Pub.'Year & ODS & OIS & AP \\
      \midrule
      \multirow{3}{*}{Traditional} 
      & gPb-UCM \cite{arbelaez2010contour} & TPAMI'10 & 0.632 & 0.661 & 0.562\\
      & SE \cite{dollar2014fast} & TPAMI'14 & 0.695 & 0.708 & 0.679\\
      & OEF \cite{hallman2015oriented} & CVPR'15 & 0.751 &0.767 & -- \\
      \midrule 
      \multirow{7}{*}{CNN} & 
      HED \cite{xie2015holistically} & ICCV'15 & 0.720 & 0.734 & 0.734\\
      & RCF \cite{liu2019richer} & CVPR'17 & 0.729 & 0.742 & -\\
      & AMH-Net \cite{xu2017learning} & NeurIPS'17 & 0.744 & 0.758 & 0.765\\
      & LPCB \cite{deng2018learning} & ECCV'18 & 0.739 & 0.754 & -\\
      & BDCN \cite{he2019bi}& CVPR'19 & 0.748 & 0.763 & \textbf{\textcolor{blue}{0.770}}\\
      & PiDiNet \cite{su2021pixel} & ICCV'21 & 0.733 & 0.747 & -\\
      & PEdger-L \cite{fu2023practical}& ACMMM'23 & 0.742 & 0.757 & -\\
      \midrule
      \multirow{4}{*}{Transformer} 
      & EDTER \cite{pu2022edter} & CVPR'22 & 0.758 & 0.771 & 0.747 \\
      & DiffEdge \cite{ye2024diffusionedge} & AAAI'24 & \textbf{\textcolor{blue}{0.761}} & \textbf{\textcolor{blue}{0.766}} & - \\
      & SED-SViT-BP & - & 0.772 & 0.789 & 0.802 \\
      & SED-SViT-LP & - & \textbf{\textcolor{red}{0.791}} & \textbf{\textcolor{red}{0.811}} & \textbf{\textcolor{red}{0.819}} \\
      \bottomrule
    \end{tabular}
\end{table}

\section{Conclusion}
In this paper, we introduce an adaptive multi-stage non-edge pruning method based on the characteristics of edge detection. This approach generates intermediate edge score maps at different Transformer layers and employs thresholds to effectively identify and remove high-confidence non-edge tokens. Furthermore, we redesign the conventionally complex decoder structure into a streamline edge detection architecture that is inherently pruning-friendly and deployment-ready. Experimental results demonstrate that our non-edge pruning strategy significantly reduces computational costs while maintaining accuracy, and the new architecture extracts clearer and more comprehensive edge features, validating the effectiveness and superiority of the proposed framework. Thus, this work demonstrates the potential of non-edge pruning for edge detection. More advanced pruning techniques are expected in the future to further enhance the pruning performance for existing strong edge detection models.

\section*{Acknowledgements}
This work is supported in part by the Fundamental Research Funds for the Central Universities (Nankai University, No. 070-63253235) and in part by NSFC (No. 62576176). The computational resources are supported by the Supercomputing Center of Nankai University (NKSC).

%
%
\bibliographystyle{splncs04}
\bibliography{main}

\end{document}